\documentclass[letterpaper, 10 pt, conference]{ieeeconf}  

\IEEEoverridecommandlockouts
\usepackage{graphicx}
\usepackage{cite}
\usepackage{soul,color}
\usepackage{amssymb}
\usepackage{amsmath}
\usepackage{subcaption}
\usepackage{enumitem}
\usepackage{hyperref}

\begin{document}

\title{Legibot: Generating Legible Motions for Service Robots\\ Using Cost-Based Local Planners} 
\author{Javad Amirian$^{1}$, Mouad Abrini$^{1}$, and Mohamed Chetouani$^{1}$%
	\thanks{$^{1}$ All authors are with Sorbonne University, CNRS, Institut des Systemes Intelligents et de Robotique, Paris, France. \{firstname.lastname\}@sorbonne-universite.fr}%
}

\maketitle
\thispagestyle{empty}
\pagestyle{empty}

\begin{abstract}
With the increasing presence of social robots in various environments and applications, there is an increasing need for these robots to exhibit socially-compliant behaviors.
Legible motion, characterized by the ability of a robot to clearly and quickly convey intentions and goals to the individuals in its vicinity, through its motion, holds significant importance in this context.
This will improve the overall user experience and acceptance of robots in human environments.
In this paper, we introduce a novel approach to incorporate legibility into local motion planning for mobile robots.
This can enable robots to generate legible motions in real-time and dynamic environments.
To demonstrate the effectiveness of our proposed methodology,
we also provide a robotic stack designed for deploying legibility-aware motion planning in a social robot,
 by integrating perception and localization components.
\end{abstract}

\section{Introduction} \label{sec:intro}
Robotic systems have transcended their traditional roles in factories and manufacturing lines, expanding into various service-oriented domains, including
healthcare \cite{miroki2023},
hospitality \cite{dinerbot2023},
and food service \cite{plato2022}.
As robots increasingly share spaces with humans, it becomes imperative for them to not only ensure the safety of the humans around them but also to comprehend and adhere to the implicit social norms that govern human interaction \cite{singamaneni2024survey}.
This imperative gives rise to the concept of socially-compliant behaviors, wherein robots are expected to exhibit behaviors that align with human expectations \cite{de2016socialrobots}.
Francis et al. \cite{francis2023principles} have identified eight distinct sets of principles that collectively define social compliance for robots, encompassing aspects such as safety, politeness, legibility, and so on.

\vspace{0.2cm}
\textbf{Legibility}, as outlined in \cite{lichtenthaler2016legibility_review}, is a dimension with significant potential for development within the realm of robotics.
Legibility in robot navigation is the capacity of a robot to convey its intentions and goals to people in its vicinity through its movement patterns clearly and promptly.
This ability is crucial, enhancing user comprehension, trust, and the overall experience with robots \cite{dautenhahn2005companion_friend, sebastian2021explainable}.
Furthermore, it facilitates smoother cooperation between humans and robots \cite{dragan2015effects}.

At times, legibility might involve a robot exaggerating its movements to ensure its actions are understood.
Alternatively, in more intricate scenarios, it might relate to a robot's adherence to social norms and exhibiting behaviors deemed appropriate for a given setting.
Central to the concept of legibility is the \textit{observer}, typically a human, with whom the robot aims to communicate its intentions \cite{chakraborti2019explicability, SLOTV2022}.
This observer may have either a complete or partial view of the surrounding environment.

\begin{figure}
    \centering
    \includegraphics[width=0.44\textwidth]{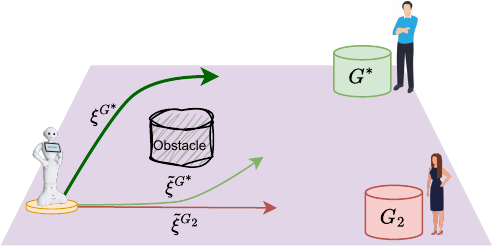}
    \caption{The robot is tasked with delivering an item to the target person $G^*$, while other people are also present in the scene $\{G_2\}$.
             A standard (non-legible) local planner would generate the light-green path $\tilde{\xi}^{G^*}$, as it minimizes the planning cost.
             The proposed legibility-aware planner, however, incorporates a similarity cost term that pushes the planner to generate a path with lower similarity to the unintended path $\tilde{\xi}^{G_2}$ (in red),
             and it generates the dark green path at the top $\xi^{G^*}$, which decreases the confusion for the observer, and improves the legibility of the robot's motion.
    }
    \label{fig:fig1}
    \vspace{-0.5cm}
\end{figure}

\vspace{0.1cm}
\noindent
In this paper, we propose the following \textbf{contributions}:

\begin{enumerate}[leftmargin=*]
    \item We introduce a novel approach to computing legible motion in cost-based local planner for mobile robots.
    Being able to re-generate the motion plan at any time, our approach does not assume complete knowledge of observers and can adapt to dynamic changes in the environment and partial observability of the observers.
    The proposed algorithm, first, predicts the observer's expectations about the robot's motion, by assuming different goal options for the robot.
    It then generates a final motion plan that maximizes similarity with the predicted motion for the actual goal,
    and minimizes similarity with the predicted motions for unintended goals, as illustrated shown Fig.~\ref{fig:fig1}.

    \item We present a robotic stack for deploying legibility-aware motion planning in real-world scenarios.  
    The proposed architecture is designed using the ROS concepts of messaging between nodes, and it can be deployed on Non-ROS systems as well.

    \item We showcase the effectiveness of our approach in a simulated restaurant scenario, where a mobile robot delivers items to tables to fulfill customer orders.
    We also conduct an offline user study to evaluate the legibility of the robot's motions from the perspective of human observers, utilizing video recordings from the experiments.
\end{enumerate}

\vspace{0.2cm}
The \textbf{structure} of the paper is as follows: In Section II, we review related work.
Section III presents our method for computing legible motions for social mobile robots.
In Section IV, we explain the experimental results and present the user study.
Finally, Section V concludes the paper and discusses limitations and future work.

\section{Related Work} \label{sec:related_work}

Interest in \textbf{human-aware navigation} within the robotics community has led to the development of socially-aware motion planners.
These planners aim to generate motions adhering to social norms, addressing the critical ability of a robot to perceive, understand, predict, and respond to human behaviors for effective navigation in human-populated environments \cite{singamaneni2023Benchmarking}.
Despite significant advancements, the challenge of ensuring the legibility of the robot's motion—its ability to clearly convey its intentions and objectives through movement to nearby individuals—remains a pressing issue with many unresolved challenges \cite{mavrogiannis2023core}.

\vspace{0.2cm}
Dragan et al. \cite{dragan2013legibility} differentiate \textbf{legibility} from predictability in human-robot collaboration, offering formal definitions and illustrating the delicate balance between the two.
Extensions of this work include considerations of the observer's viewpoint and occlusion scenarios \cite{nikolaidis2016based}, as well as observer-aware legibility for enhancing motion planning in environments with limited fields of view \cite{taylor2022observeraware}.
Faria et al. \cite{faria2021understanding} introduce a method for legible motion planning in multi-party settings, emphasizing the need to account for the perspectives of all human participants in the interaction.
Bastarache et al. \cite{bastarache2023multiagent_legible} tackle legibility in scenarios with multiple moving agents, while Mavrogiannis et al. \cite{mavrogiannis2018social} focus on estimating and communicating intended avoidance behaviors to enhance motion expressiveness.
A study on observer-aware legibility \cite{wallkotter2022EvalLegibility} underscores the importance of considering the observer's environment in planning,
particularly in applications like robot-assisted services in restaurants, showing that personalized paths increase goal inference accuracy but may reduce performance for unintended observers.

\vspace{0.2cm}
\textbf{Learning-based} methods for legible robot navigation have also been explored.
SLOT-V \cite{SLOTV2022} utilize human-labeled robot trajectories to learn observer models, while Bronars et al. \cite{bronars2023_GLMM} propose an end-to-end framework for learning legible trajectories from multi-modal human demonstrations.
However, the reliance on labeled datasets of legible trajectories is not always practical.
Reinforcement learning (RL) strategies \cite{busch2017learning, zhao2020actor_critic_legible}, including the use of an imaginary observer for feedback in the learning process \cite{Manubied2020IRLLegibility}, demonstrate the potential of RL in generating legible robot motions.
However, these solutions are yet to generalize to new real-world scenarios.

\vspace{0.2cm}
Exploring other \textbf{non-verbal modalities} such as signaling lights in vehicles, enhances communication of intentions to surrounding individuals \cite{angelopoulos2022non_verbal}.
Dugas et al. \cite{dugas2020_IAN} and Shrestha et al. \cite{shrestha2016light_and_display} investigate the effectiveness of gesture, nudging and auditory signals in improving social navigation,
while gestures have also been recognized as a means to enhance legibility in human-robot interaction \cite{lichtenthaler2013legibility_review_old}.
Comparatively, Hetherington et al. \cite{hetherington2021hey} and Angelopoulos et al. \cite{angelopoulos2022acoustic_cues} explore the use of projected arrows, flashing lights, and acoustic signals to communicate robot intentions,
though these methods may not be universally applicable across all robot platforms and environments.

\section{Method} \label{sec:method}

\subsection{Problem Statement}

We assume the task of a service robot delivering items to a human target, denoted by $\mathrm{G}^*$, within a dynamic environment.
The robot's \textit{actual} goal together with other \textit{unintended} goals $\mathrm{G}^i$, form the set of
robot's potential goals:

\begin{equation}
    \label{eq:goals_set}
    \mathcal{G} = \{\mathrm{G}^*\} \cup \{\mathrm{G}^i\}_{i=2}^{N}
\end{equation}

\noindent
where $N$ is the number of all the potential goals, each of which is a 2D point $\mathrm{G}^i = (x^i, y^i)$.
In the environment, there will be a non-empty set of human observers $\mathcal{O} = \{O^j\}_{j=1}^{M}$ that are interested in the robot's behavior, where $M$ is the number of observers.
An observer $O^j$ is denoted by a 3D vector $O^j = (x^j, y^j, \theta^j)$, where $(x^j, y^j)$ is the 2D position of the observer, and $\theta^j$ is the heading orientation of the observer.
They are also associated with a field of view parameter $\theta^j_{\text{fov}}$.
The robot's objective is to efficiently reach its goal in a way that is clear to the observers,
meaning its movements should be straightforward for the observers to interpret, ensuring the robot's goal is predictable as it progresses.
It is important to note that in this scenario, both sets $\mathcal{O}$ and $\mathcal{G}$ refer to the individuals detected around the robot, though generally, they may differ.
For instance, the robot's task could involve approaching an object while observers monitor its actions.

Finally, the obstacle-free space of the robot's environment $\mathcal{E}$, is denoted by $\mathcal{E}_{free}$, which might change over time.
We are interested in finding a path $\xi_{t_0:t_w}$ from the current time $t_0$ to the time $t_w$ that is legible for the target observer,
while satisfying the robot's task, i.e. reaching its goal, avoiding collision with static and dynamic obstacles,
moving within the robot's kinematic and dynamic constraints, and so on.
In practice, $\xi_{t_0:t_w}$ is modeled as a sequence of 2D waypoints, with a fixed time interval between them,
so we use the discrete notation $\xi_{0:w} = \{q_0, q_1, \ldots, q_w\}$, where $q_i = (x^i, y^i)$ is the robot's waypoint at time $t_0 + i \Delta t$
and $q_0$ is the current position of the robot, also denoted by $s$.

\subsection{Cost-based Local Planning}

A local motion planner generates a trajectory $\xi_{0:w}$ based on a set of cost functions.
Choosing the time window $w$ ensures that, in the worst-case scenario, the robot can stop to avoid collisions, given its kinematic and dynamic properties.
Simultaneously, this window is minimized to allow for quick computation of new plans in response to changes in the dynamic environment.
Our cost function, $C_{\text{Task}}(\xi)$, comprises a weighted sum of several factors: the distance to the goal, the closest distance to any obstacles, the rate of approaching an obstacle,
the smoothness of the path, and the robot's velocity.
This list of factors is not exhaustive, and additional elements can be incorporated into the cost function as needed.
Our legibility-aware planner builds upon this local planner, with further modifications detailed below.

\subsection{Legibility-Aware Local Planning}

To generate intent-expressive motion plans for the robot, we first hypothesize the observers' beliefs about the robot's trajectory for a given goal,
represented as $\tilde{\xi}^G_{0:w}$, where $\tilde{}$ indicates an estimated local path for each goal $G \in \mathcal{G}$.
These estimations are obtained through a local planner by optimizing the task-specific cost function $C_{\text{Task}}(\xi)$ for each goal $G$.

In order to generate legible paths, we introduce a similarity cost function $C_{\text{Sim}}(\xi_{0:w})$ that quantifies the likeness between the given trajectory $\xi_{0:w}$ and the predicted trajectories $\tilde{\xi}^G_{0:w}$ for each goal $G \in \mathcal{G}$.
This cost function penalizes similarity to the trajectories of unintended goals while encouraging similarity to the trajectory of the robot's true intended goal $G^*$:

\begin{equation}
    \label{eq:cost_sim}
    C_{\text{Sim}}(\xi_{0:w}) = \left[ \sum_{G \in \mathcal{G} \setminus G^*} \text{sim}\left(\xi_{0:w}, \tilde{\xi}^G_{0:w}\right) \right]
                                    - \text{sim} \left(\xi_{0:w}, \tilde{\xi}^{G^*}_{0:w}\right)
\end{equation}

\noindent
where $\text{sim}(.,.)$ denotes a similarity function that quantifies the likeness between two trajectories.
We utilize the cosine similarity function, which compares the velocity vectors of the trajectories at each time step (denoted by $\dot{\xi}_t$), as follows:
\begin{equation}
    \label{eq:cos_similarity}
    \text{sim}(\xi^A_{0:w}, \xi^B_{0:w}) = \sum_{t=0}^{w} \frac{\dot{\xi}^A_t \cdot \dot{\xi}^B_t}{\| \dot{\xi}^A_t \| \| \dot{\xi}^B_t \|}
\end{equation}

The choice of cosine similarity between the velocity vectors is motivated by its robustness to scale and its ability to capture the directional alignment between two motion trajectories.

\subsection{Observer Perspective}\label{subsec:observer-perspective}
In scenarios where a service robot is tasked with delivering objects to a person, understanding the person's field of view is crucial for addressing legibility from the observer's standpoint.
We propose modifications to the formulation by Dragan et al. \cite{dragan2013legibility} to better suit these conditions:

\vspace{0.2cm}
\noindent
1) We move away from the assumption that the observer continuously observes and understands the robot's entire trajectory.
Unlike in \cite{dragan2013legibility}, where the robot's trajectory is optimized for observer legibility throughout, using a weighting function $f(t) \propto \frac{1}{t}$ to highlight early goal communication,
our approach considers the possibility that the observer may not perceive parts of the trajectory.
Instead, we propose addressing legibility in a localized manner,
using a weighting function $h(q) \propto \frac{d(q, G^*)}{d(q, G)}$ to prioritize clearer communication of the goal when the robot is farther from its actual goal $G^*$ compared to other goals $G \in \mathcal{G} \setminus G^*$.
Here, $d(.)$ denotes the Euclidean distance.

\vspace{0.2cm}
\noindent
2) Recognizing that the observer has a finite field of view angle, we refine our legibility cost function to omit trajectory segments outside the observer's field of view.
We introduce a boolean visibility function $v(q)$ that returns true if the robot's position $q$ is within the observer's field of view $O^j$:

\begin{equation}
    \label{eq:fov}
    v(q) = \begin{cases}
        1 & \text{if } \theta_{\text{dev}} \leq \theta_{\text{fov}}/2 \\
        0 & \text{otherwise}
    \end{cases}
\end{equation}
where $\theta_\text{dev}$ measures the deviation angle between the observer's central view and the robot's position at time $t$:

\begin{equation}
    \label{eq:theta_dev}
    \theta_\text{dev} = \arccos\left(\frac{q - O^*}{\|q - O^*\|} \cdot u(\theta^*)\right)
\end{equation}

\noindent
and $u(.)$ is the unit vector function.

We incorporate the visibility function $v(q)$ and the weighting function $h(q)$ into the similarity function to derive the legibility-aware similarity function:

\begin{equation}
    \label{eq:sim_modified}
    \text{sim}(\xi_{0:w}, \tilde{\xi}^G_{0:w}) = \sum_{t=0}^{w} v(q_t) h(q_t)
                        \frac{\dot{\xi}_t \cdot \dot{\tilde{\xi}}^{G}_t}{\| \dot{\xi}_t \| \| \dot{\tilde{\xi}}^{G}_t \|}
\end{equation}

This equation updates the $\text{sim}$ function in Eq. \ref{eq:cost_sim} for calculating the similarity cost $C_{\text{Sim}}$.

\noindent
3) We introduce a cost function, inspired by Taylor et al. \cite{taylor2022observeraware}, designed to encourage the robot to remain within the observer's field of view, denoted $C_{\text{FOV}}(\xi)$.
This function aims to promote the robot's navigation within the observer's visible area:

\begin{equation}
    \label{eq:fov_cost}
    C_{\text{FOV}}(\xi) = \sum_{t=0}^{w} \text{tanh}\left(\frac{\theta_\text{dev}(t)}{\theta_\text{fov}/2}\right)
\end{equation}

\noindent
Using the hyperbolic tangent function ensures a smooth transition in cost, maxing out at 1,
which encourages the robot to stay within the observer's field of view, minimizing penalties for positions close to the visibility boundary.

The comprehensive Legibility-Aware cost function integrates the task-specific cost,
the similarity cost as detailed in Eq. \ref{eq:cost_sim},
and the field of view cost outlined in Eq. \ref{eq:fov_cost}.
It is formulated as:

\begin{equation}
    \label{eq:cost_legibility_aware}
    C_{\text{LA}}(\xi) = C_{\text{Task}}(\xi) + \lambda_{\text{sim}} C_{\text{Sim}}(\xi) + \lambda_{\text{fov}} C_{\text{FOV}}(\xi)
\end{equation}
where $\lambda_{\text{sim}}$ and $\lambda_{\text{fov}}$ serve as the respective weighting coefficients for the similarity and field of view costs.

\begin{figure}
    \centering
    \subfloat{\includegraphics[width=0.24\textwidth, trim={3.2cm 1.92cm 2.5cm 2.1cm}, clip]{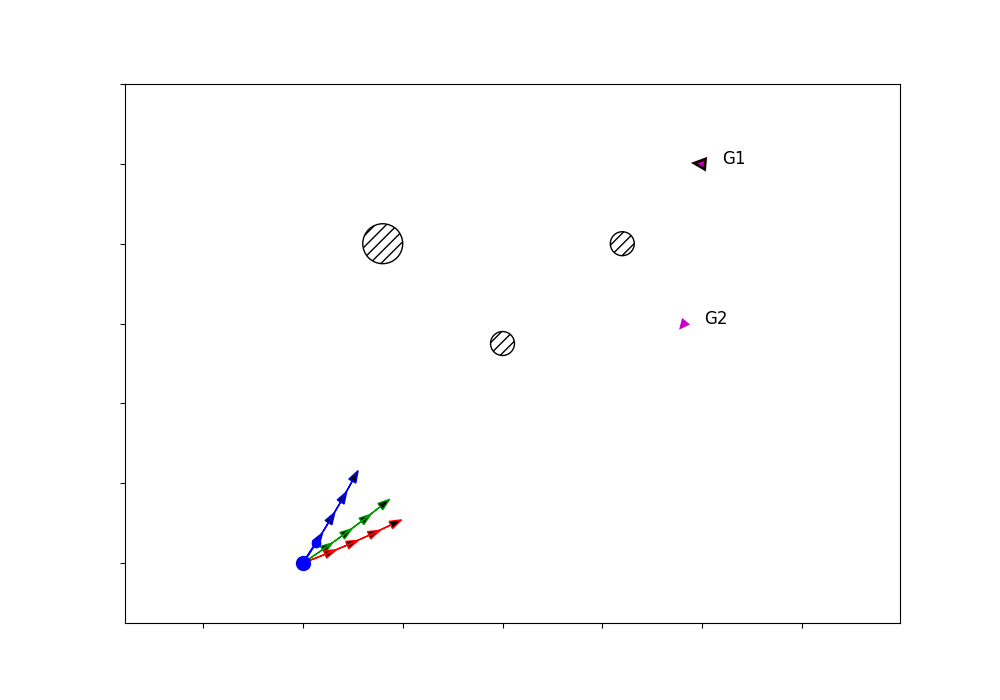}}
    \hfill
    \subfloat{\includegraphics[width=0.24\textwidth, trim={3.2cm 1.92cm 2.5cm 2.1cm}, clip]{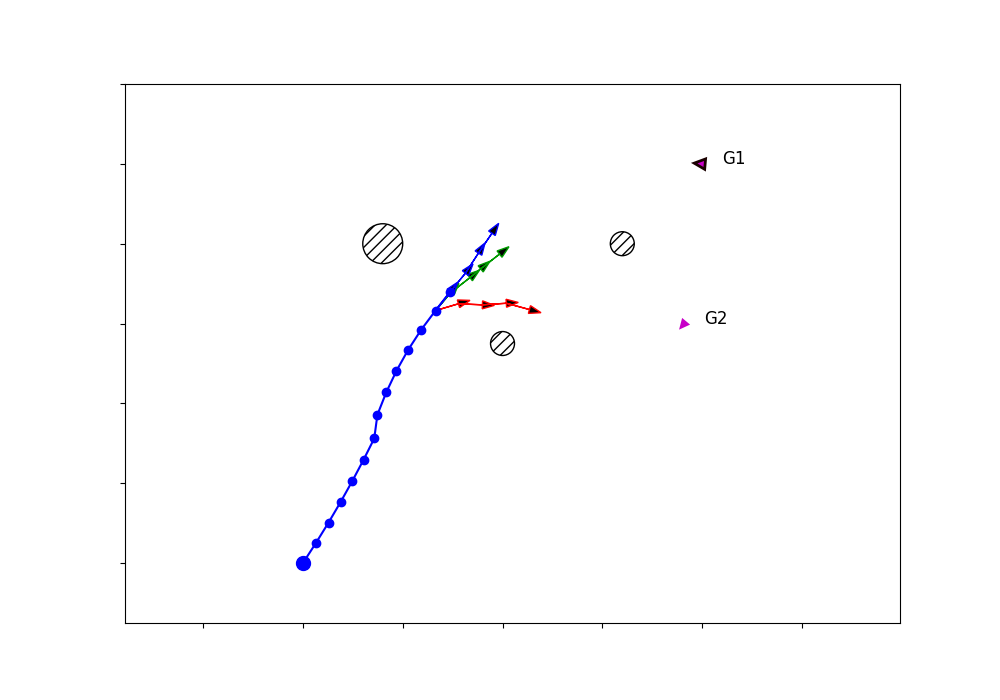}}
    \hfill
    \subfloat{\includegraphics[width=0.24\textwidth, trim={3.2cm 1.92cm 2.5cm 2.1cm}, clip]{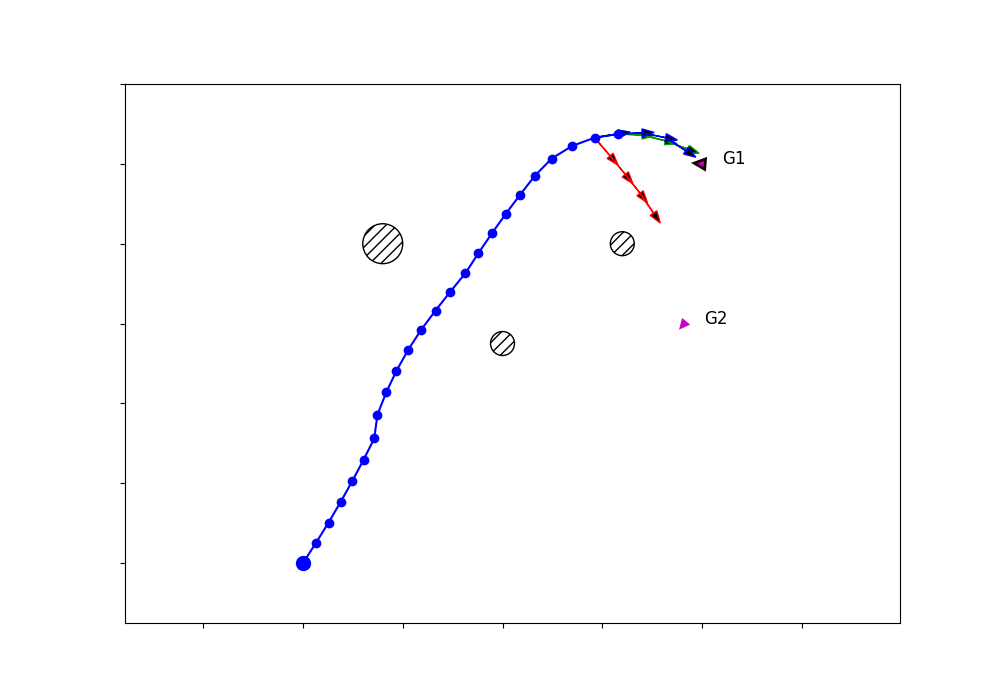}}
    \hfill
    \subfloat{\includegraphics[width=0.16\textwidth]{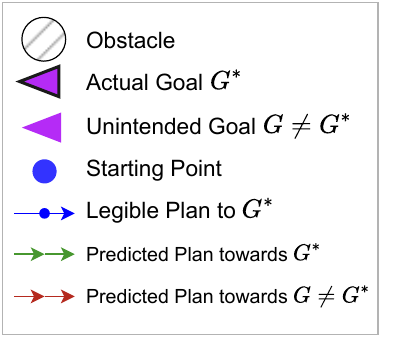}}
    \caption{Generating legible paths using the proposed legibility-aware motion planning algorithm.
    The sequence of images illustrates the planner's progression towards the goal $G_1$ at time steps 0, 7.2$s$, and 13.2$s$, respectively.
    Red and green arrows show the pre-computed local path for goals $G_1$ and $G_2$, respectively.
    }
    \label{fig:legible_paths_toy}
    \vspace{-0.3cm}
\end{figure}

Fig. \ref{fig:legible_paths_toy} illustrates the effectiveness of our legibility-aware motion planning approach with a straightforward example.
We define two goals: $G_1$ as the robot's actual target and $G_2$ as an alternative, unintended destination.
Initially, the robot computes potential paths towards both goals, depicted in green and red.
Subsequently, using the proposed legibility-aware cost function (Eq. \ref{eq:cost_legibility_aware}), the robot adjusts its route to improve legibility towards the actual goal,
as shown in the initial step (top-left corner) of Fig. \ref{fig:legible_paths_toy}.
This adjustment, guided by the similarity cost $C_{\text{Sim}}$ (Eq. \ref{eq:cost_sim}), ensures the path modification better matches the actual goal,
demonstrating the algorithm's capacity to prioritize legibility in its navigation strategy.

\begin{figure}
    \centering
    \subfloat
    {\includegraphics[width=0.2\textwidth, trim={4.8cm 2.9cm 3.5cm 3.1cm}, clip]{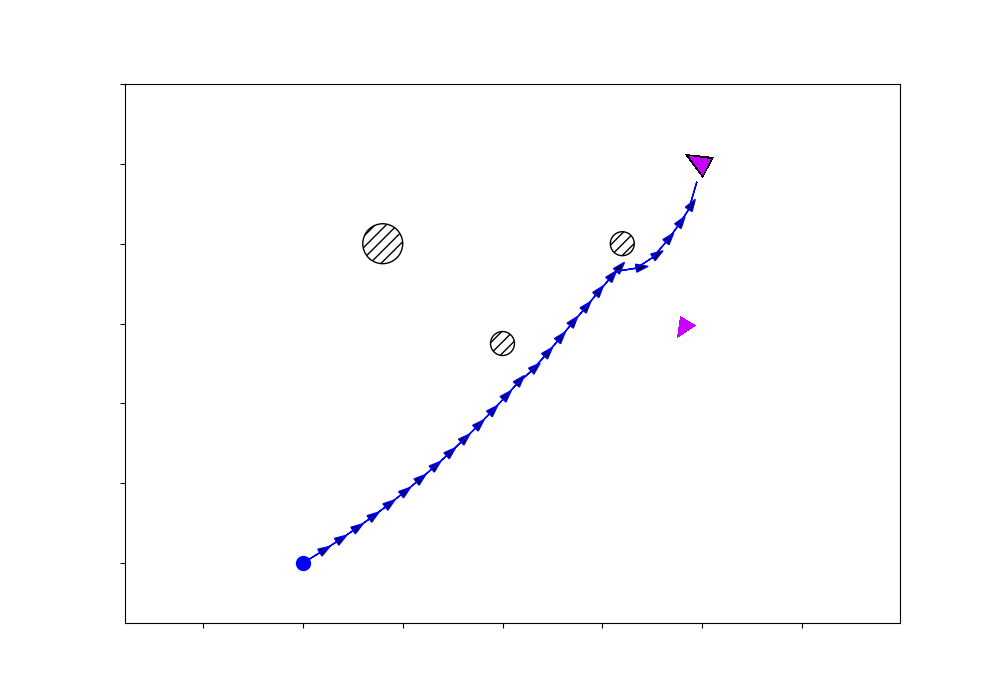}
    \label{fig:legible_plan}
    }
    \subfloat
    {\includegraphics[width=0.2\textwidth, trim={4.8cm 2.9cm 3.7cm 3.1cm}, clip]{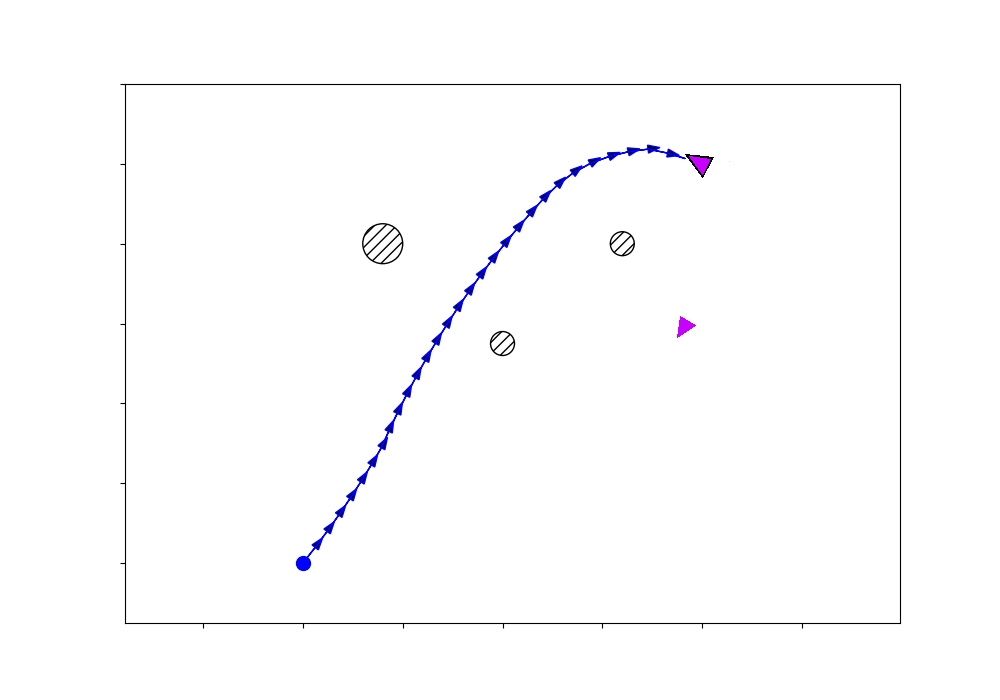}
    \label{fig:illegible_plan}
    } \hfill
    \caption{Comparing the plan generated by the standard planner with cost function $C_{\text{Task}}$ (left)
             against the output of the legibility-aware motion planning algorithm (right).
    }
    \label{fig:toy_example_legible_vs_illegible}
    \vspace{-0.3cm}
\end{figure}

Fig. \ref{fig:toy_example_legible_vs_illegible} showcases a comparison between legible and illegible trajectories as generated by our legibility-aware algorithm versus traditional motion planning approaches.
In the depicted illegible scenario, the robot navigates around obstacles from the right, inadvertently aligning itself with a non-target goal. This misdirection may lead to confusion among observers, highlighting the importance of legibility in path planning.

\begin{figure}
    \centering
    \subfloat[]
    {\includegraphics[width=0.14\textwidth, trim={4.8cm 2.9cm 3.7cm 3.1cm}, clip]{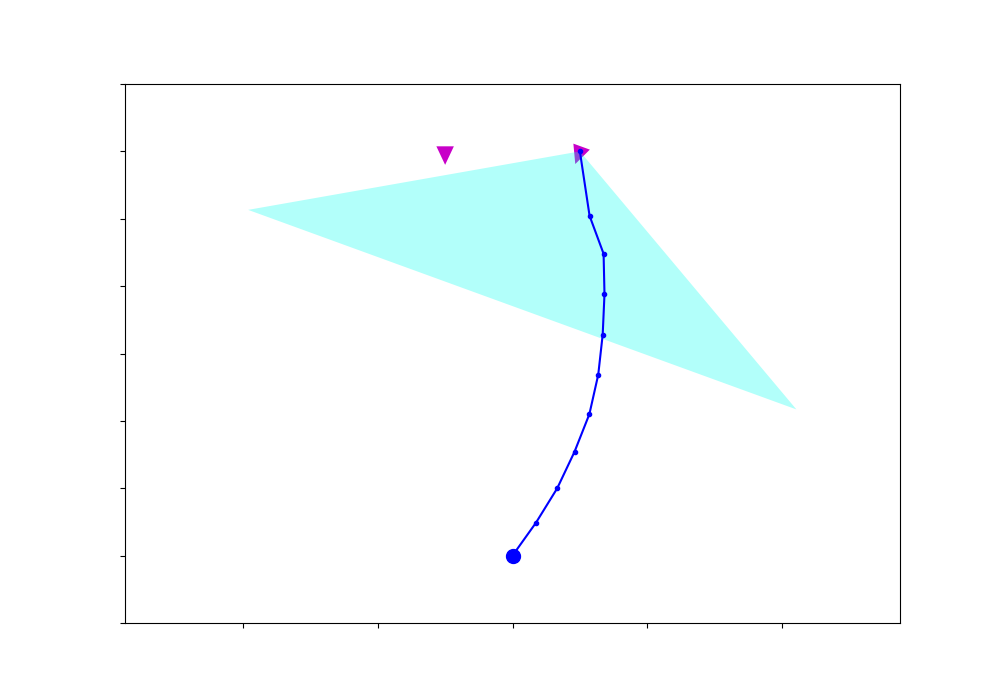}
    \label{fig:fov110}
    } \hfill
    \subfloat[]
    {\includegraphics[width=0.14\textwidth, trim={4.8cm 2.9cm 3.7cm 3.1cm}, clip]{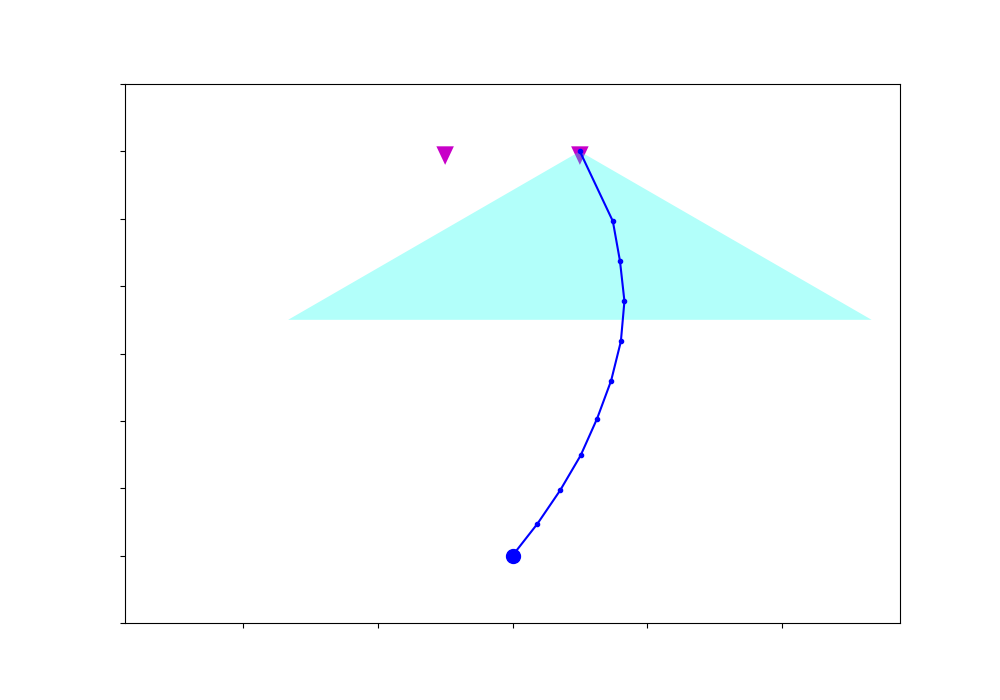}
    \label{fig:fov90}
    } \hfill
    \subfloat[]
    {\includegraphics[width=0.14\textwidth, trim={4.8cm 2.9cm 3.7cm 3.1cm}, clip]{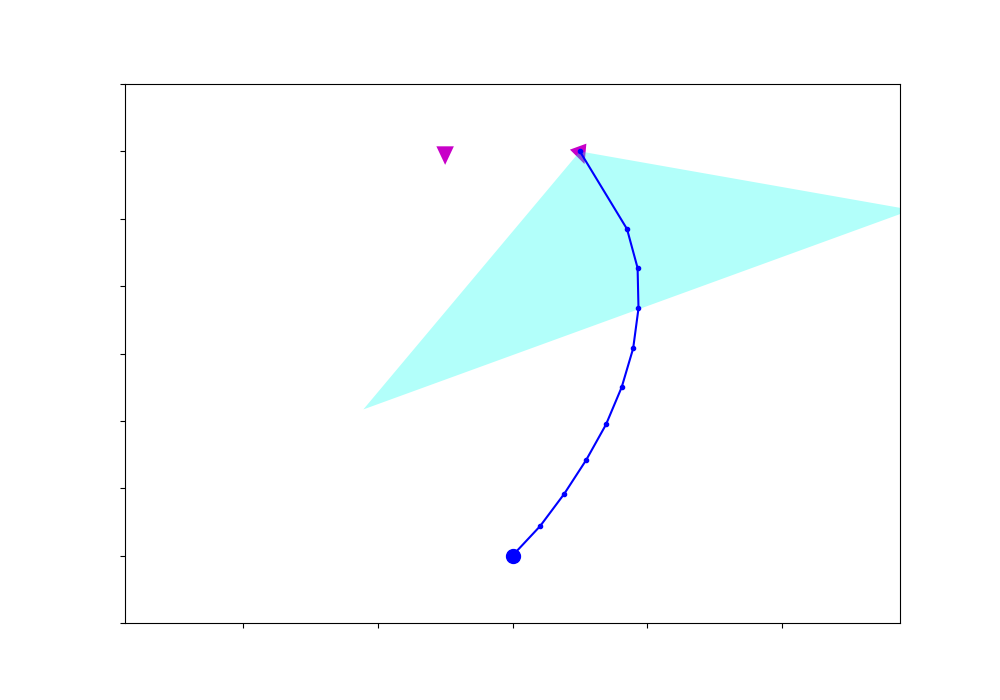}
    \label{fig:fov70}
    } \hfill
    \caption{
    The effect of the field of view cost function $C_{\text{FOV}}$ on the motion plan generated by the legibility-aware planner.
    The triangles in cyan represent the field of view of the observer, while assuming their depth is infinite.
    }
    \label{fig:C_fov}
    \vspace{-0.3cm}
\end{figure}

Also, in Fig. \ref{fig:C_fov}, we illustrate the effect of the field of view cost function $C_{\text{FOV}}$ on the robot's path planning.
As we can see, when the observer turns their head from right (Fig. \ref{fig:fov110}) to the center (Fig. \ref{fig:fov90}) and then to the left (Fig. \ref{fig:fov70}),
the robot's path is adjusted to improve the robot's visibility to the observer.

\subsection{Legibility-Aware Robot Navigation Stack}
\begin{figure*}
    \centering
    \includegraphics[width=0.68\textwidth]{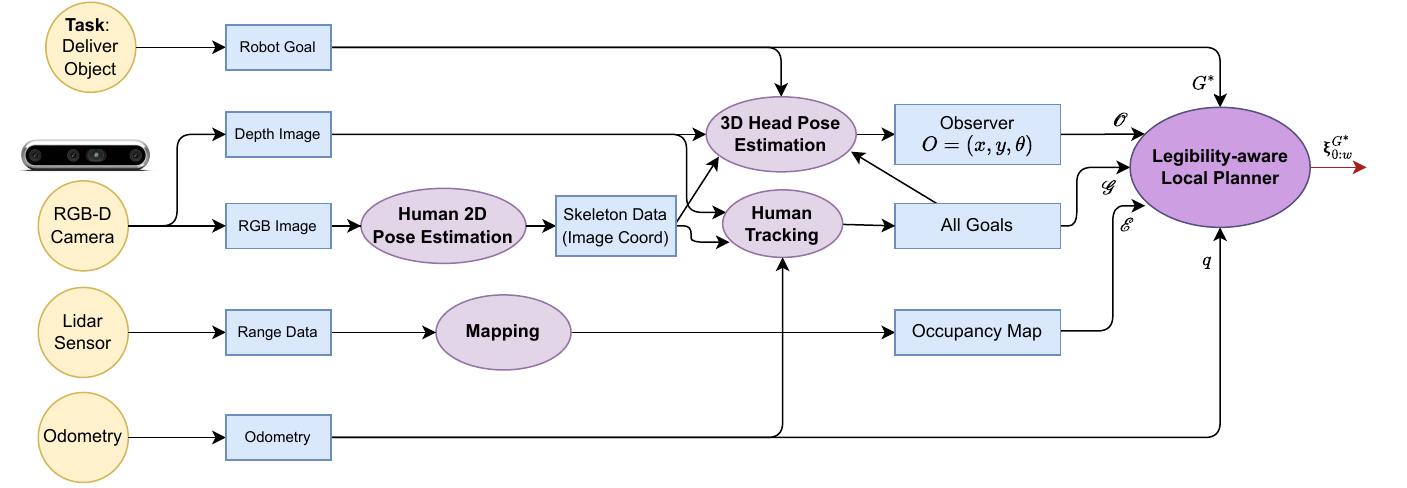}

    \caption{The diagram of the robot stack proposed for Legibility-Aware navigation of mobile robots.
    Nodes and topics are shown by ellipses and rectangles, respectively, and input devices are shown by circles.
    }
    \label{fig:legibot_stack}
\end{figure*}

To deploy our legibility-aware motion planning algorithm on a real-world robot, we introduce an integrated system architecture
illustrated in Fig. \ref{fig:legibot_stack}.
This architecture, built on the Robot Operating System (ROS) framework, ensures the required inputs for the legibility-aware planner are provided by the robot's perception system.

\begin{itemize}[leftmargin=*]
    \item An RGB-D camera serves as the primary sensor for human detection around the robot.
          This RGB image is processed by a 2D Human Pose Estimation node (YOLO), and the skeleton data are sent to the Tracking module.
          The tracker uses the depth data from the RGB-D camera on the leg joints of the detected humans to estimate their world-coordinate positions
          and maintain the continuity of the detected humans.
          The output is published as the detected humans' positions, which represent the potential goals $\mathcal{G}$.

    \item A 3D Head Pose Estimation module receives the robot's target goal $G^*$ and the tracked humans to match the target person with the detected humans
          using nearest neighbor matching.
          It then uses the head joint position from the skeleton data of the corresponding human and combines it with the depth data to estimate the observer's head pose.
          The output is published as the observer's position and heading orientation $O$.

    \item The occupancy grid map required for the planner is generated by the robot's LIDAR sensor, and the robot's odometry data are used to estimate the robot's position and velocity.
\end{itemize}

These inputs are then processed by the legibility-aware local planner, which generates the legible path $\xi_{0:w}$ for the robot to follow.
This integration enables the dynamic generation of navigation paths that are not only efficient but also easily interpretable by surrounding humans, ensuring the robot's movements are coherent and predictable from an observer's perspective.
We assume a fixed Field of View (FOV) angle for the observer ($\theta_{\text{fov}} = 120$ degrees, as suggested in \cite{taylor2022observeraware}),
accounting for the head movement of the observer.

\section{Experimental Results} \label{sec:exp}
\subsection{Simulation of Robots in Restaurant Scenarios}
The restaurant industry presents a promising domain for service robots, given their need for client interaction while adhering to social norms and expectations.
We aim to validate our proposed methodology in this context, where robots are tasked with delivering items to customers amidst other patrons.
The key challenge is ensuring legible behavior, enhancing customer confidence in the robot's intentions.

\begin{figure}
    \centering
    \subfloat[Bird view of the scene]{\includegraphics[width=0.23\textwidth]{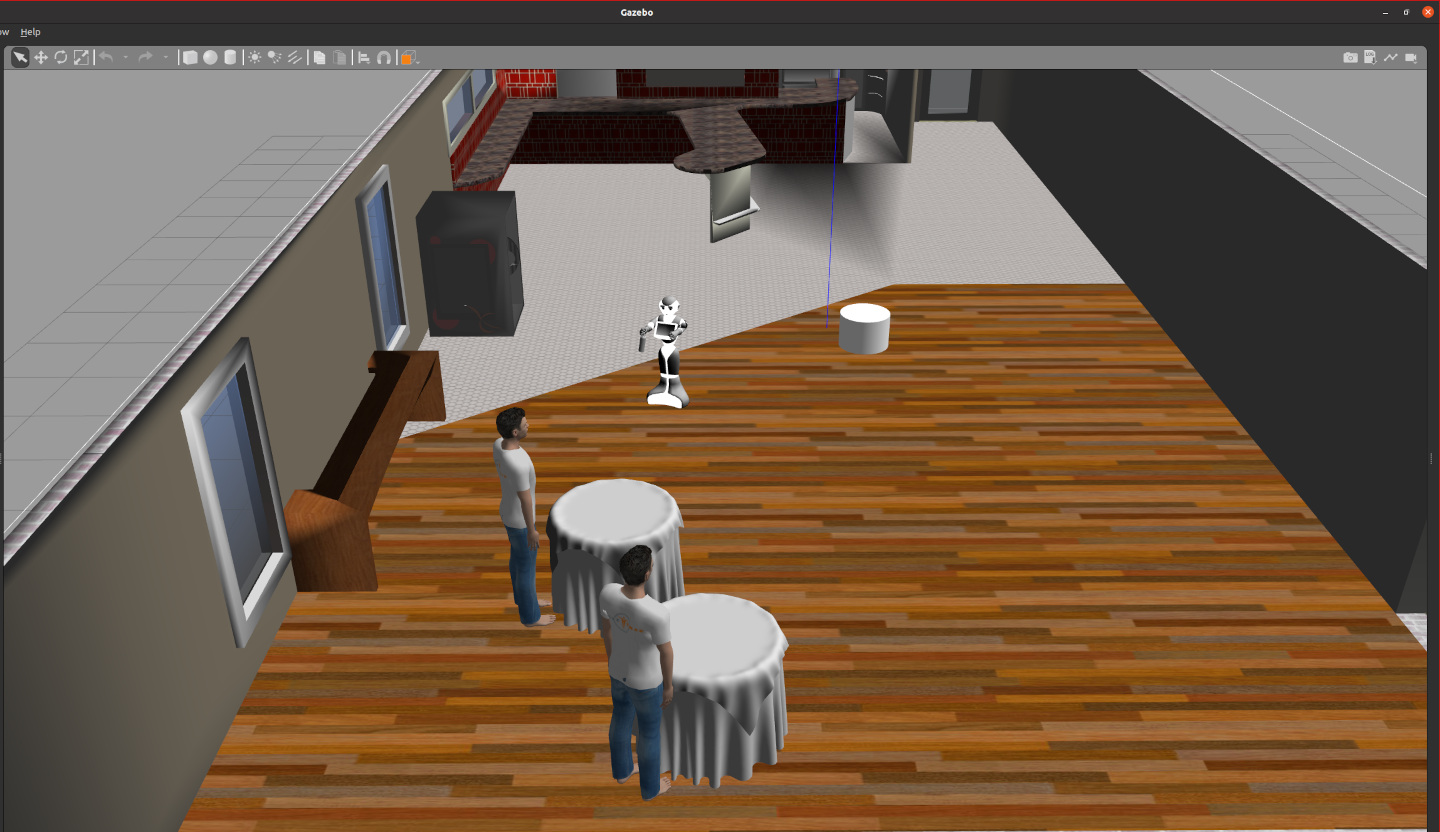}\label{fig:gazebo_bird_view}}
    \hfill
    \subfloat[Robot view]{\includegraphics[width=0.245\textwidth]{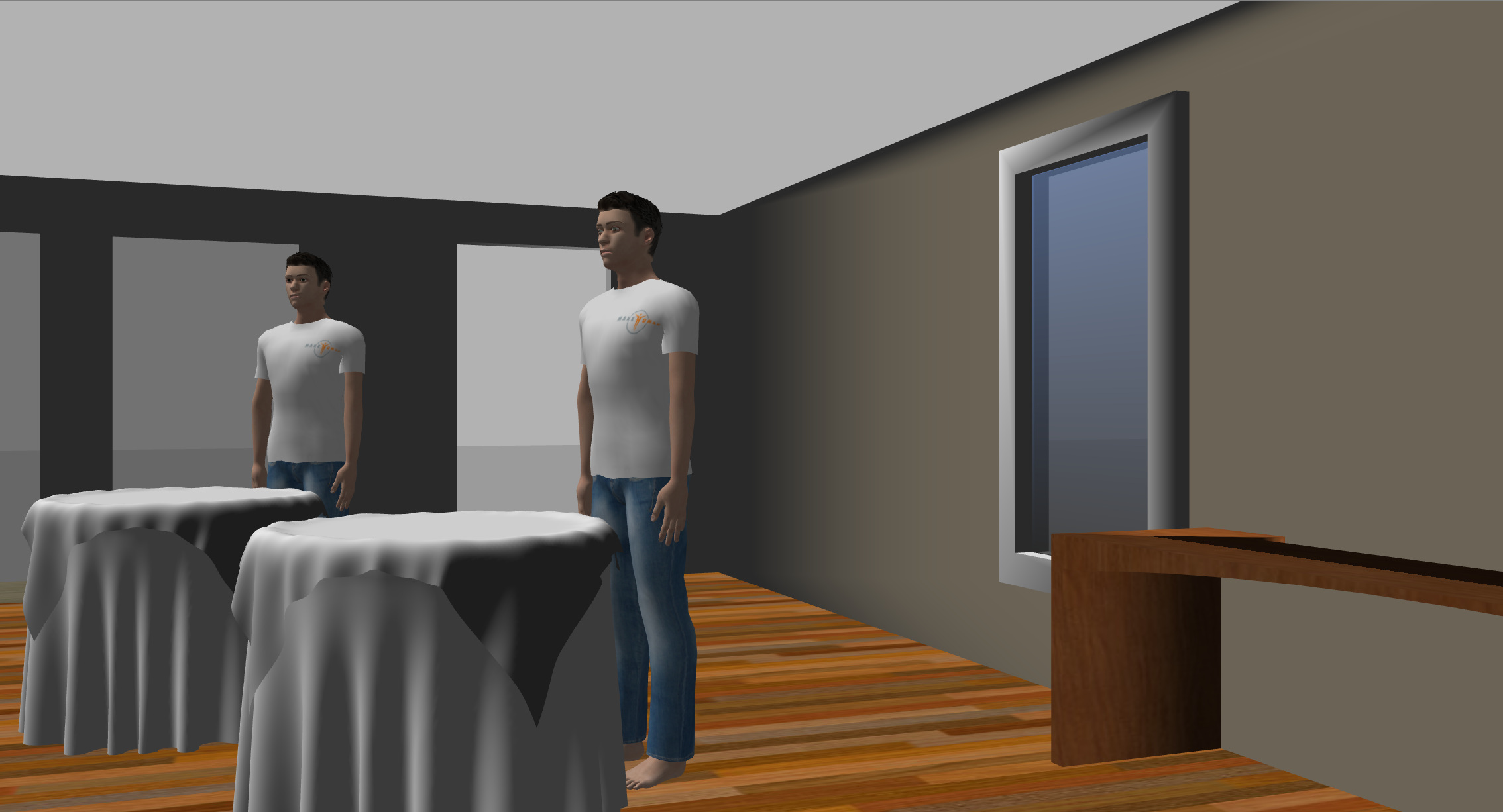}\label{fig:gazebo_robot_view}}
    \hfill
    \subfloat[Target observer view]{\includegraphics[width=0.24\textwidth]{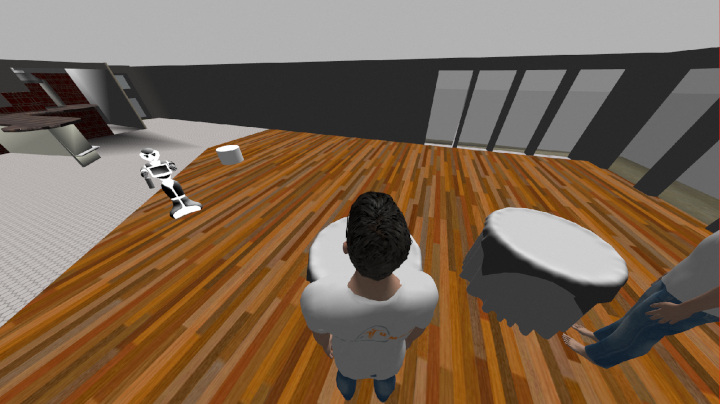}\label{fig:gazebo_observer_view}}
    \hfill
    \subfloat[Unintended observer view]{\includegraphics[width=0.24\textwidth]{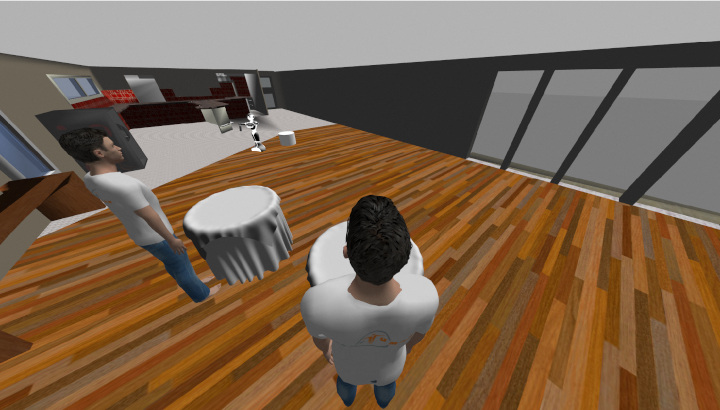}\label{fig:gazebo_unintended_view}}
    \caption{Simulation of a robot in a restaurant scenario}
    \label{fig:legible_sim_gazebo}
\end{figure}

\vspace{0.2cm}
To test our approach, we constructed a simulation environment in Gazebo.
Utilizing the Pepper robot model, we simulated the delivery of an object to a designated table within the scene.
Multiple tables are present, with one identified as the target table $G^*$.
The robot's objective is to deliver the item to this table while maintaining legibility for other individuals in the scene.
In Fig.~\ref{fig:legible_sim_gazebo}, we present the bird's eye view of the scene, and shots from the robot's and observer's perspectives.

\subsection{Hypotheses}

In order to evaluate the effectiveness of our proposed algorithm in a user study, we formulated the following hypotheses:

\begin{enumerate}[label=\textbf{H\arabic*:}]
    \item The proposed legibility-aware motion planning algorithm will significantly enhance the legibility of the robot's behavior for the target observer, compared to scenarios without legibility constraints.
    \item The proposed legibility-aware motion planning algorithm will yield a higher correctness ratio for at least one partial, compared to the non-legible planner.
    \item When isolated, the proposed algorithm will yield a high legibility score.
\end{enumerate}

\subsection{User Study}
To assess the legibility of the robot's movements in a simulation environment and evaluate our hypotheses, we conducted a comprehensive user study.

\paragraph*{\textbf{Scenario}}
We developed two simulation scenarios, each featuring two individuals: one designated as the target observer and the other as an unintended observer.
In the first scenario, both observers are positioned to face the robot directly at equal distances.
Conversely, in the second scenario, the robot approaches the target observer from a side angle, with the observers positioned at varying distances from the robot's initial position.
For each scenario, we generated two robot paths: one using our legibility-aware motion planning algorithm and a baseline path derived from solving the $C_{\text{Task}}$ cost function.
We captured video footage of the robot's motion from the perspectives of both observers and invited participants to view these recordings and assess the robot's behavior.

\begin{figure}
    \centering
    \includegraphics[width=0.4\textwidth]{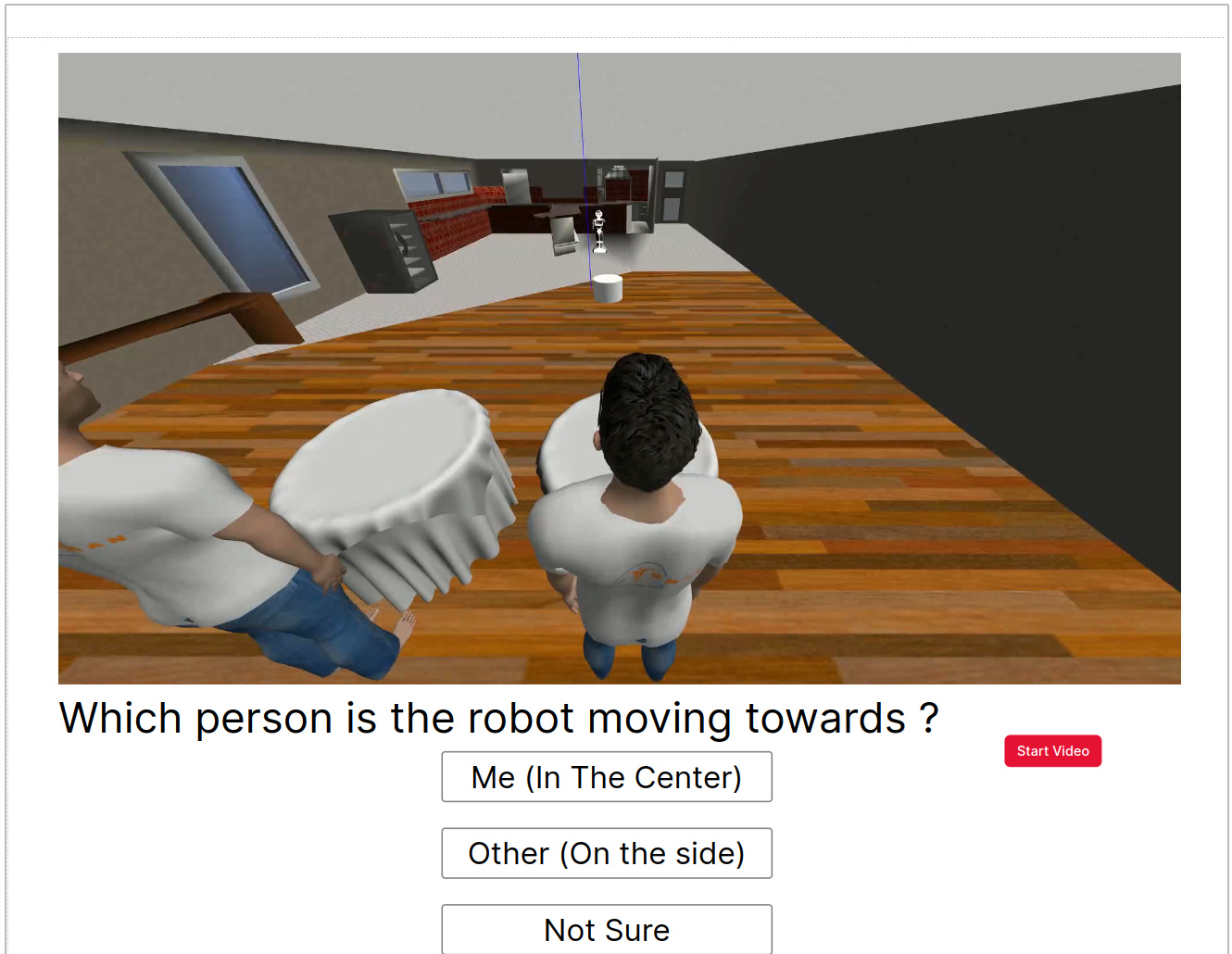}
    \caption{User study comparing the legibility of the legibility-aware planner against the baseline planner}
    \label{fig:user_study_question}
    \vspace{-0.5cm}
\end{figure}

\paragraph*{\textbf{Questionnaire}}
Participants were presented with a question: "Which person is the robot moving towards?"
They were offered three response options: "Me (In the Center)", "The person on the side", and "Not sure".
Figure~\ref{fig:user_study_question} displays a screenshot of the questionnaire.

For the user study, paths were created using both the legibility-aware planning algorithm and the baseline approach, which lacks legibility constraints, by solving the $C_{\text{Task}}$ cost function.
We then segmented each video recording into three parts (0-25\%, 25-50\%, and 50-75\% of the robot's path) and asked participants to specify which person they believed the robot was targeting in each segment.
The user study was conducted using the Gorilla Experiment Builder, with participants viewing each video segment and responding accordingly.

We enlisted 132 participants via the Prolific platform for this study.
The participant demographic comprised 33 women and 99 men, with an average age of 42 years and a standard deviation of 15 years.

\subsection{Experiment design}
\paragraph*{\textbf{Design}}
We adopted a mixed-design approach in which each participant was exposed to one trajectory from a specific planner (between-subjects factor) and observed all partials of that trajectory (within-subjects factor).
Our design included 2 between-subjects factors (with and without legibility constraints) and 3 within-subjects factors (partials of the trajectory).
This method was chosen to avoid biasing participants by exposing them to trajectories from both planning approaches.

To assess \textbf{H1}, we developed a metric that integrates the within-subjects factors through a weighted mean,
giving greater importance to the initial segments of the trajectory. It is expressed in Eq. \ref{eq:legibility} as follows:
 \begin{equation}
    \label{eq:legibility}
    L = \frac{1}{\sum_{k=1}^{n}\frac{1}{k}}\sum_{k=1}^{n} \frac{1}{k} c_k
 \end{equation}
Here, $L$ denotes the legibility score, $n$ represents the total number of partials (3 in this context), $\frac{1}{k}$ specifies the weight of the $k$-th partial, and $c_k$ signifies the correctness rate of the $k$-th partial.
\paragraph*{\textbf{A-Priori Power Analysis}}
To calculate the necessary sample size, we conducted an a-priori power analysis using G*Power software for a Wilcoxon-Mann-Whitney test. We set the effect size to 0.5 (medium effect), the alpha level to 0.05, and the power to 0.8.
This analysis indicated a need for 106 participants to detect a potential significant difference between the two planners. Adding a 25\% margin, we aimed for a total of 132 participants.
A non-parametric test was chosen based on a pilot study indicating our data would not follow a normal distribution.

We planned to conduct three pairwise comparisons between the partials of the trajectory to evaluate \textbf{H2}.
We did not plan any comparisons within conditions, as our interest lay solely in between-condition differences.

For \textbf{H3}, we assumed that a highly legible trajectory would score between 0.8 and 1.0 on average.
Thus, we aimed to compute a confidence interval for the legibility score of the legibility-aware planner to determine if it fell within this anticipated range.
\paragraph*{\textbf{Subject Allocation}}
For each group, we designed two scenarios: one where the robot's goal is the observer, and another where the robot's goal is a different human, as depicted in Fig \ref{fig:legible_sim_gazebo}.
Participants were randomly assigned to one of the two groups and then to one of the scenarios.
To enhance the study's robustness, each scenario was set up in two different arrangements.
Concerns about bias arising from participants consistently choosing the same goal across arrangements were mitigated by having only two trials, each randomly assigned.

\subsection{Results and Discussion}
\begin{figure}
    \centering
    \includegraphics[width=0.4\textwidth]{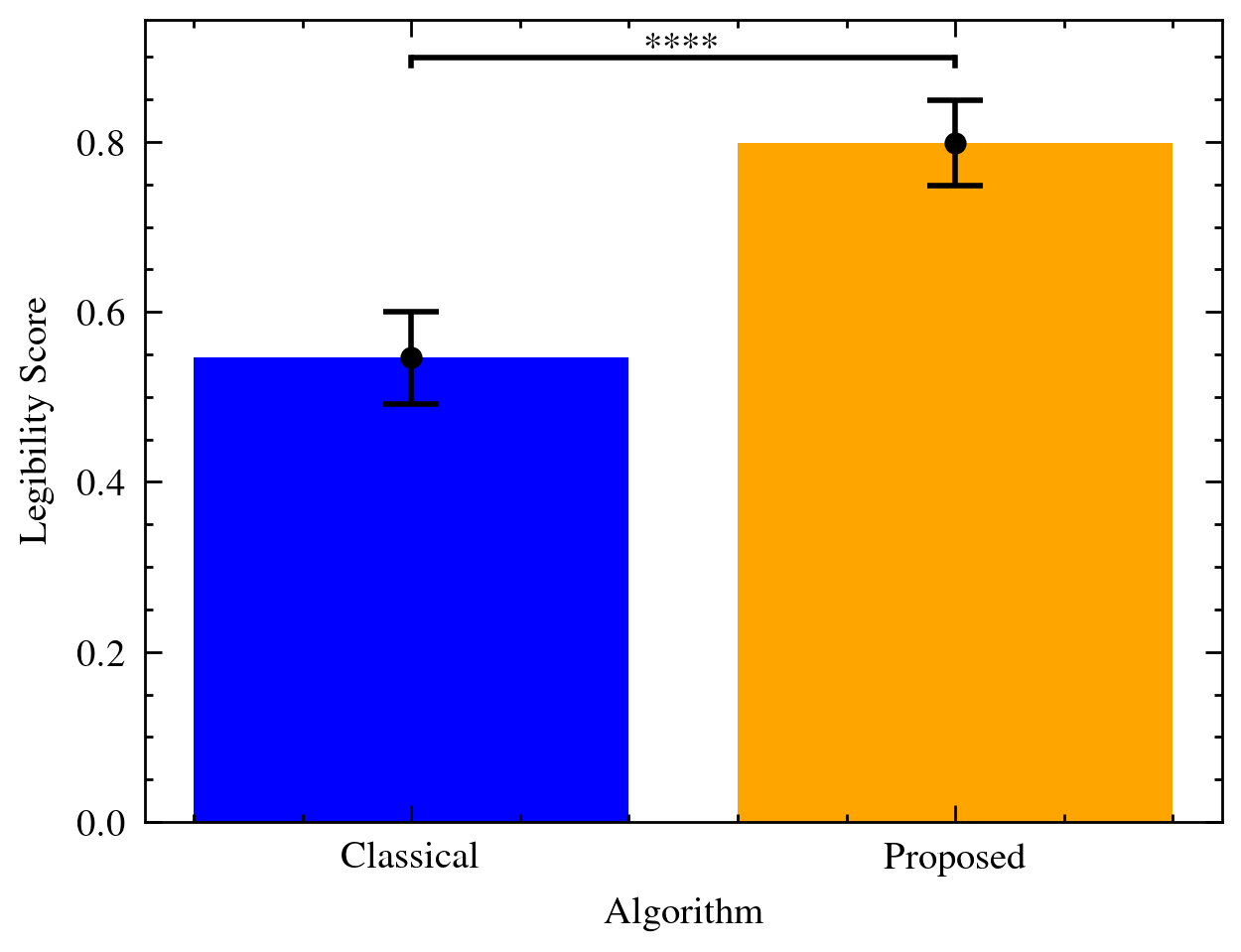}
    \caption{Legibility scores for the two planners, with confidence intervals, computed using 1000 resamples.}
    \label{fig:legibility_scores}
\end{figure}

The Mann Whitney U test showed that the legible planner had a significantly higher legibility score than the non-legible planner ($U = 3417, p < 0.001$), which supports \textbf{H1}. 
Fig \ref{fig:legibility_scores} shows the means and confidence intervals of the legibility scores for the two planners.

\begin{figure}
    \centering
    \includegraphics[width=0.46\textwidth]{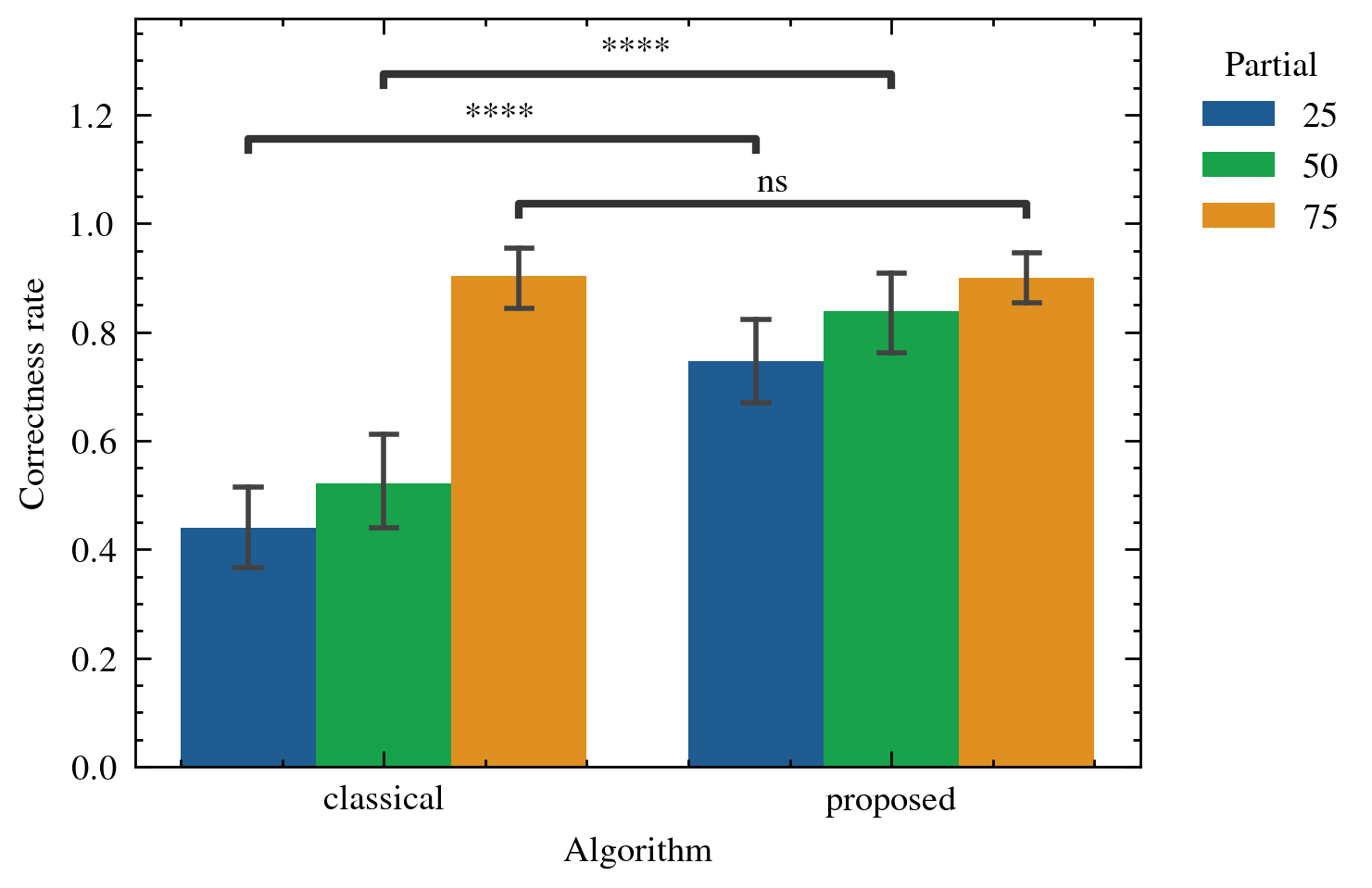}
    \caption{Correctness ratios for the partials of the trajectory, with confidence intervals and significance levels.}
    \label{fig:correctness_ratios}
    \vspace{-0.5cm}
\end{figure}

The pairwise comparisons with Benjamini-Hocheberg correction showed that the first and second partials had a significantly higher correctness ratio than that of the classical planner ($p < 0.0001$ for both), but the third partial did not ($p = 0.53$), which supports \textbf{H2}. This is represented in Fig \ref{fig:correctness_ratios}.
This aligns with the computed legibility score for \textbf{H1}, as the first and second partials are assigned a higher weight in calculating the legibility score.\\
However, even though the legible planner achieved a significantly higher legibility score, it cannot be deemed highly legible.
It may perform better relatively, but both planners might not reach the desired level of legibility.
This insight prompted \textbf{H3}, leading us to calculate a confidence interval through bootstrapping with 1000 resamples for the legibility score of the legible planner.
The interval was $[0.75, 0.85]$, partially meeting the anticipated range of $[0.8, 1.0]$. This suggests that while the trajectories generated by the legible planner are legible, they do not achieve high legibility,
unlike the classical planner, whose confidence interval was $[0.5, 0.6]$, significantly below the legibility threshold.

\subsection{Experiments with Real Robot}

We conducted experiments with a real robot to demonstrate the effectiveness of our proposed methodology
in a lab environment, using the Pepper robot, which is widely recognized in the service robotics domain.
To enhance the robot's perception and computation capabilities, we upgraded its hardware and software stack.
For details on the hardware modifications, please refer to the Appendix.

\begin{figure}
    \centering
    \subfloat[]{\includegraphics[width=0.125\textwidth]{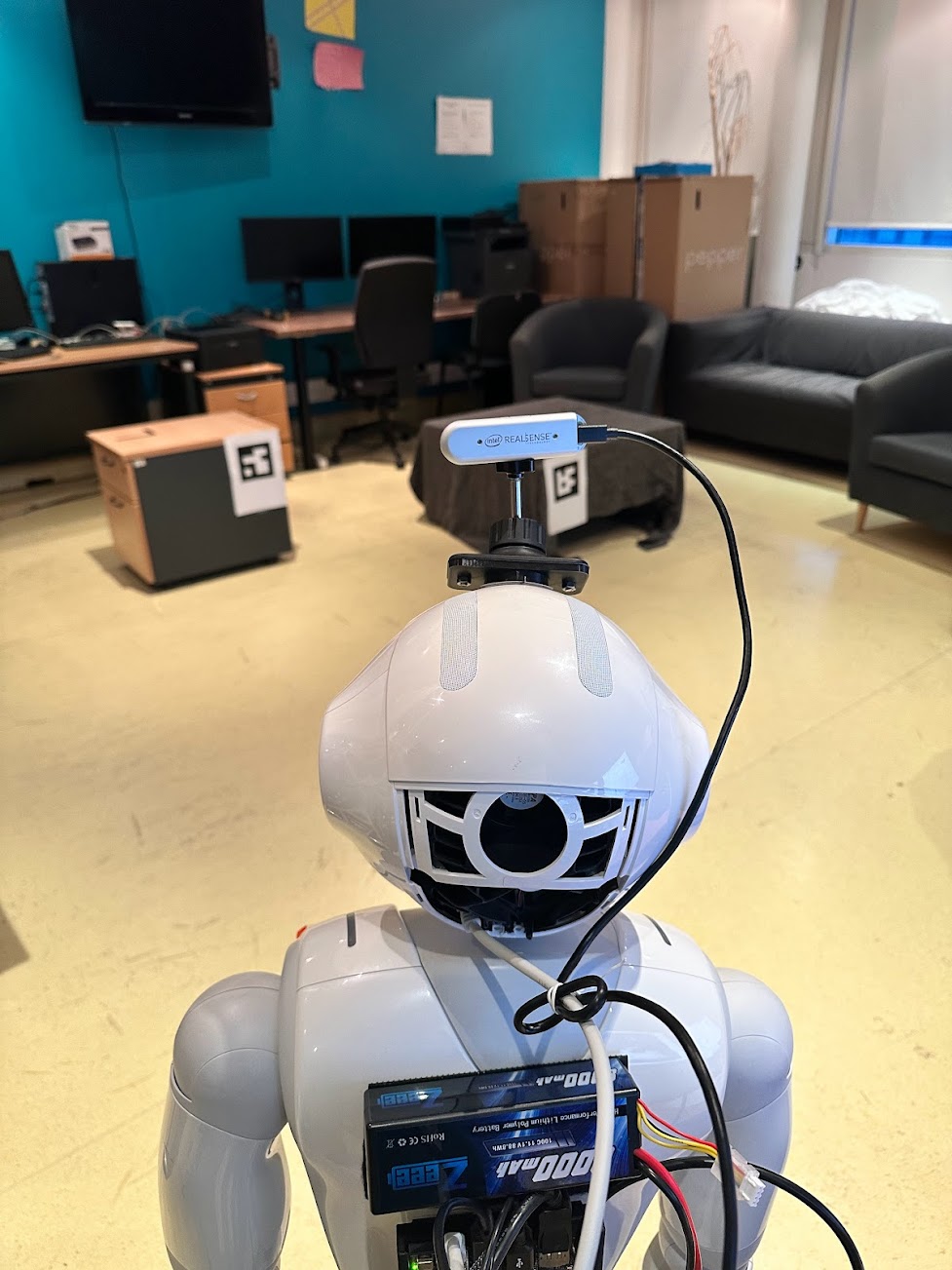}}
    \hfill
    \subfloat[]{\includegraphics[trim={11cm 0 0 0}, clip, width=0.225\textwidth]{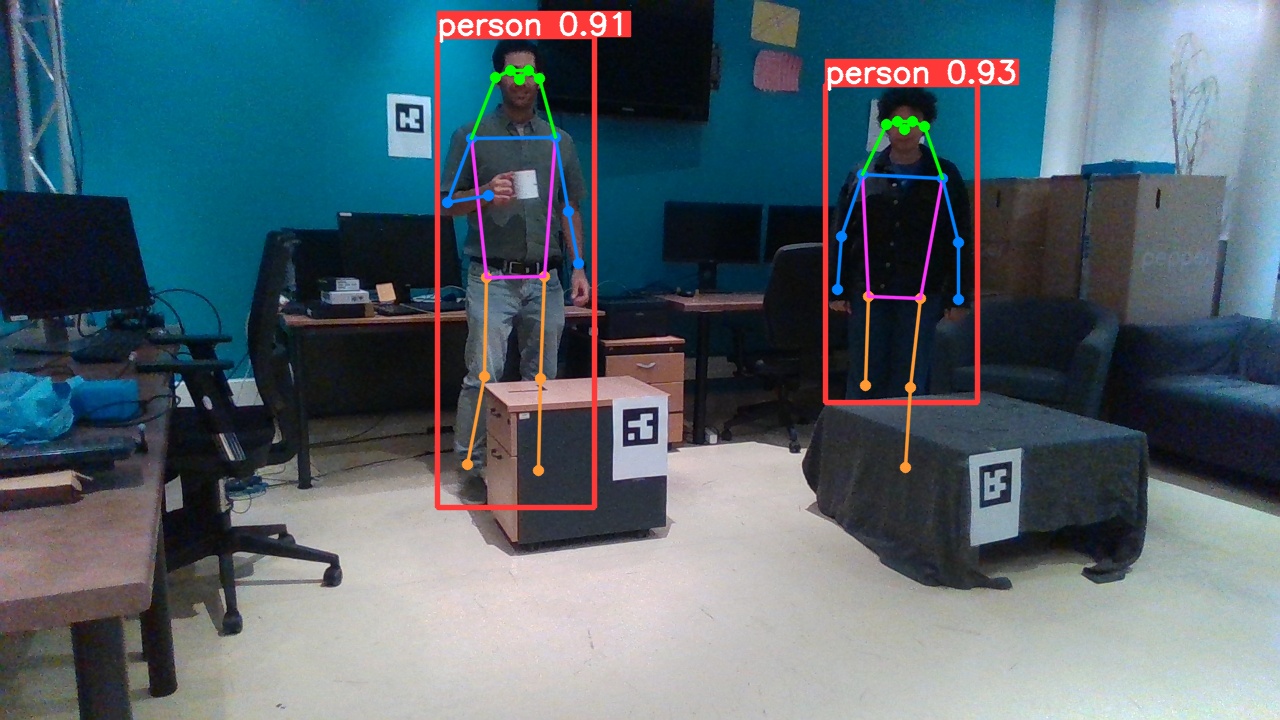}}
    \hfill
    \subfloat[]{\includegraphics[trim={3.5cm 6cm 3cm 4cm}, clip, width=0.125\textwidth]{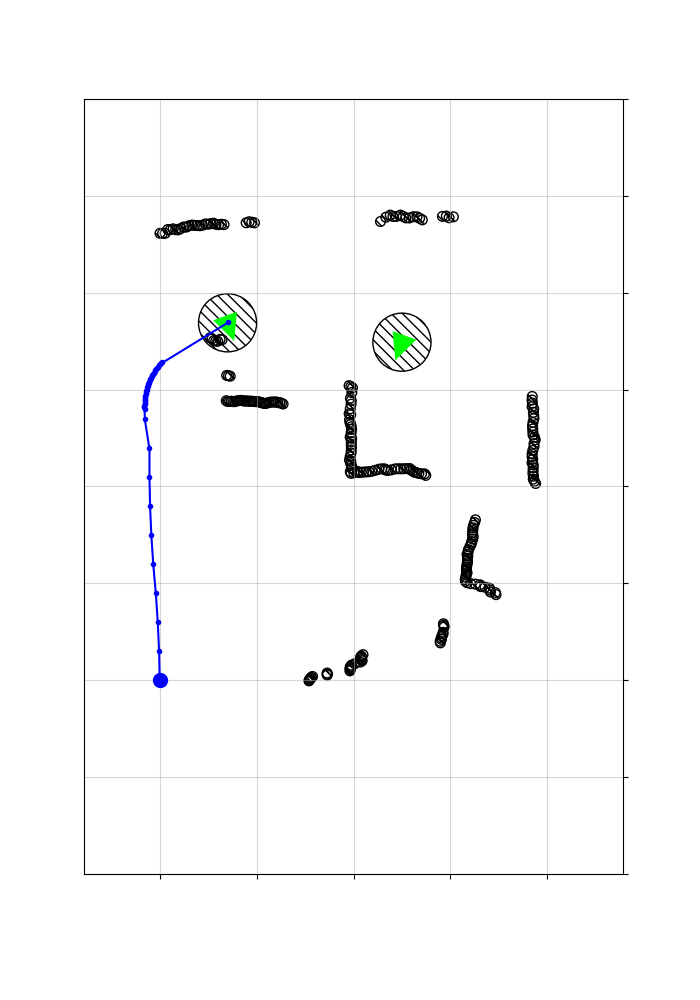}}
    \caption{Experiment with real Pepper robot.
             (a) Overview of the scene layout from the robot's perspective,
             (b) Visualization of the detected people in the scene, perceived by the robot's realsense camera,
             (c) The obstacle map perceived by the 2D lidar sensor (black circles), the detected persons with their orientations (green triangles) and the legible path planned by the robot (blue) starting from the blue marker going towards the target person (on the left).
    }
    \label{fig:real-exp}
    \vspace{-0.5cm}
\end{figure}

On the software side, we utilized ROS2 Foxy as the middleware and developed the stack outlined in Section~\ref{sec:method} to implement the legible motion planning algorithm on the robot.
To integrate this stack with Pepper, we employed the NAOqi ROS2 bridge, enabling the exchange of ROS messages with the robot and interaction with Naoqi (version 2.9).
For human detection in the scene, we processed RGB-D images from the robot's camera using the YOLOv8 ultralytics model, specifically the YOLOv8m-pose version.
This model offers an optimal balance between speed and accuracy, achieving a mAP of 88.8\% on the COCO dataset \cite{UltralyticsYOLO}.
Fig.~\ref{fig:real-exp} illustrates an experiment conducted with the real Pepper robot in our lab environment.

\section{Conclusion} \label{sec:conclusion}
In conclusion, Legibot presents a novel approach to enhancing the legibility of robot motion in service-oriented environments.
Through integrating a legibility-aware local planning mechanism within the robot navigation stack,
this work addresses the critical challenge of enabling robots to effectively communicate their intentions to human observers through motion
and improve the overall user experience and social acceptance of service robots.

By simulating a service robot in a restaurant environment, and designing two simple scenarios of robot delivery tasks, and finally conducting a user study,
we showed that the proposed legibility-aware local planner can significantly improve the expressiveness of robot motion, and enhance the ability of human observers to predict the robot's goals.
We also demonstrated the feasibility of deploying the proposed approach in real-world scenarios, by integrating it into a robotic stack for a Pepper robot, and conducting experiments in a lab setting.

We believe that there is a significant potential for future work in this area, to improve the robustness and generalizability of the proposed approach, and to further evaluate its effectiveness in more complex and dynamic environments.
We acknowledge the challenge of finding a balance between the robot's task efficiency and the legibility of its motion, (i.e., setting the coefficients $\lambda_{\text{sim}}$ and $\lambda_{\text{fov}}$ in Eq. \ref{eq:cost_legibility_aware}).
and this problem remains an open question for future research.
This becomes particularly important in situations where the similarity cost encourages a zig-zag motion to avoid similarity with unintended paths, which may not be desirable in practice,
and can be improved by incorporating additional constraints or objectives in the motion planning problem.

While the study acknowledges its limitations and outlines avenues for future research, it marks a substantial step forward in the development of socially-aware robot navigation systems.
This research not only addresses a key challenge in human-robot interaction but also opens up new possibilities for the deployment of service robots in complex, dynamic environments where communication through motion is paramount.

\section*{APPENDIX}
\begin{figure}
    \centering
    \subfloat
    {\includegraphics[width=0.16\textwidth, trim={0cm 0cm 0cm 0cm}, clip]{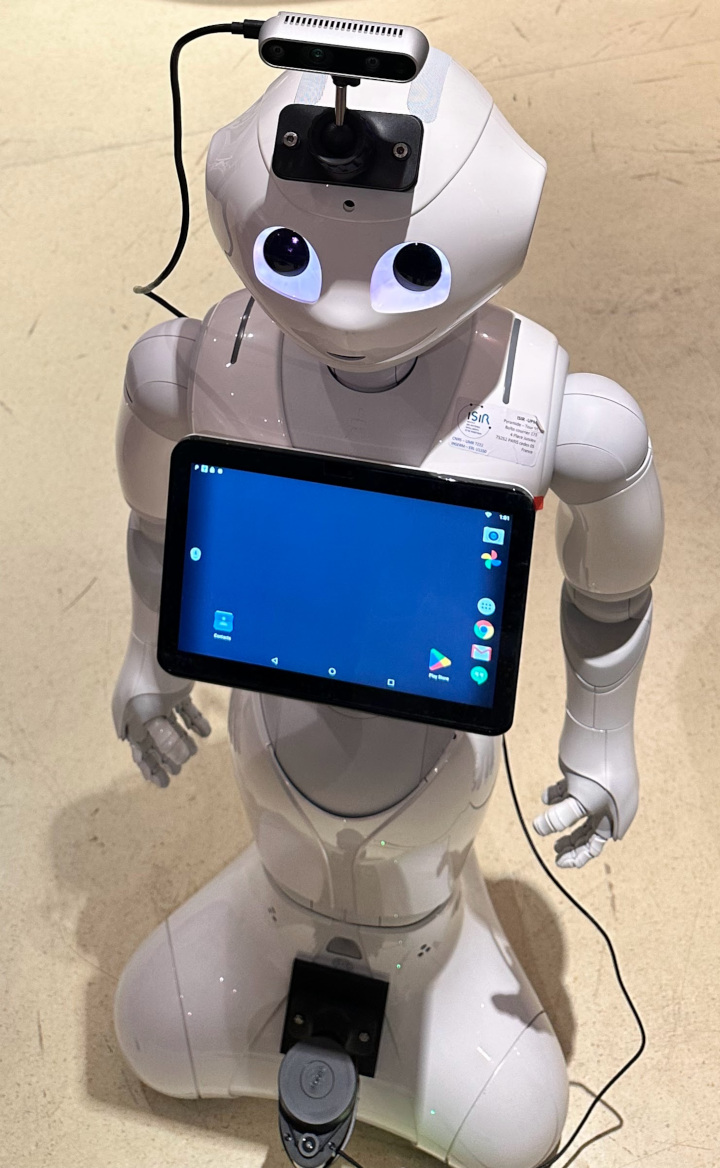}
    \label{fig:p1}
    }
    \subfloat
    {\includegraphics[width=0.16\textwidth, trim={0cm 2cm 0cm 0cm}, clip]{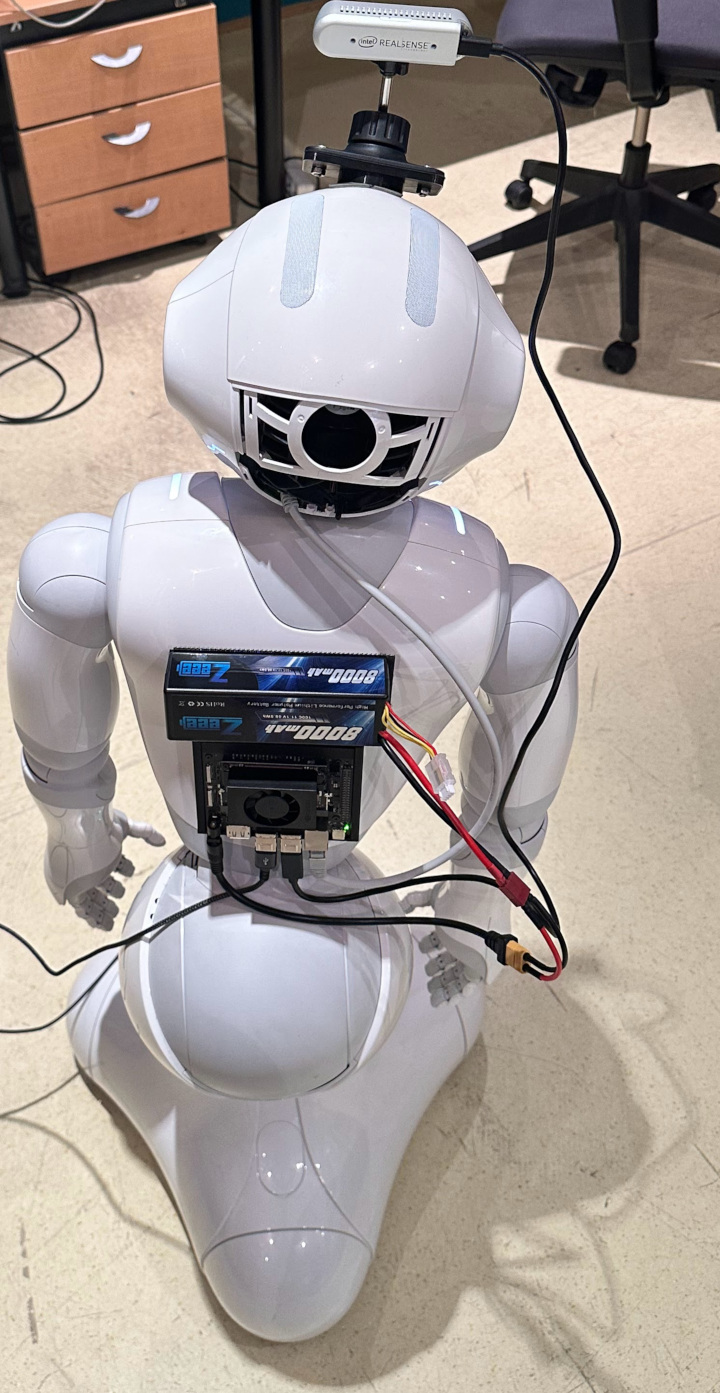}
    \label{fig:p2}
    }
    \subfloat
    {\includegraphics[width=0.16\textwidth, trim={0cm 0cm 0.7cm 0cm}, clip]{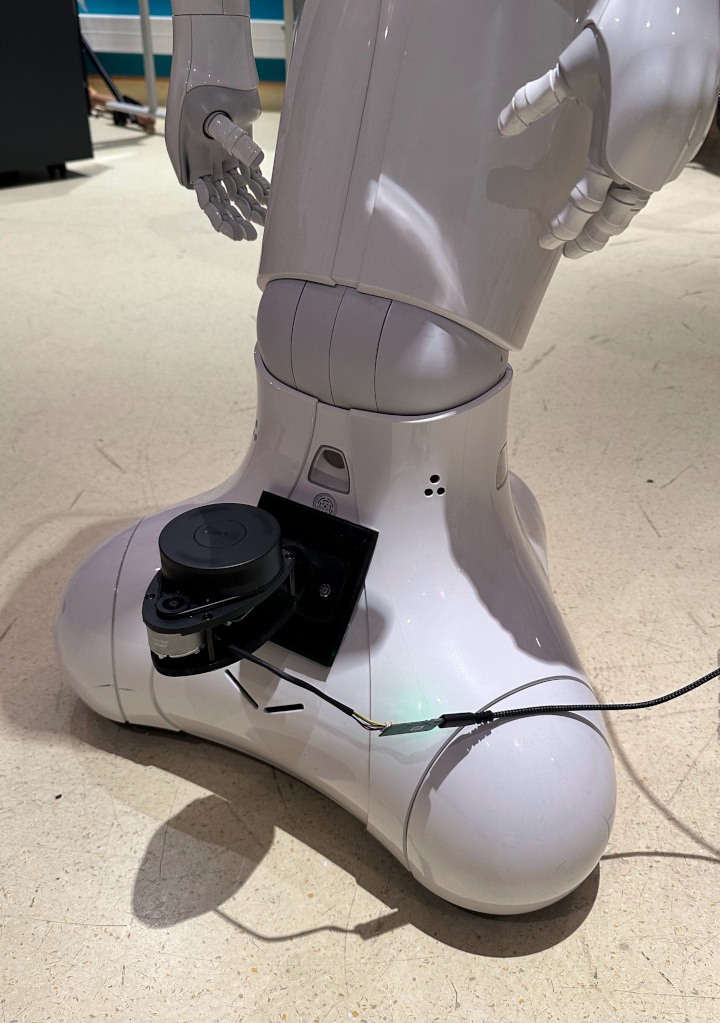}
     \label{fig:legend}
    }
    \caption{Modified Pepper robot}
    \label{fig:Pepper-modified}
    \vspace{-0.5cm}
\end{figure}

The Pepper robot's original hardware lacked the necessary power to support our proposed methodology's computational requirements \cite{caniot2020adapted_pepper}.
Specifically, its Intel Atom processor fell short in processing capacity for real-time execution of the perception and planning algorithms.
Additionally, the robot's camera offered low resolution and low frame rate for capturing the environment, and it lacked a depth sensor for perceiving the structure of the environment.
To address these limitations, we implemented several hardware upgrades to the Pepper robot:

\begin{itemize}[leftmargin=*]
    \item We added a Intel RealSense D435i RGB-D camera to the robot's head, which is a high-resolution camera with a depth sensor.
          The RGB sensor in the camera has a resolution of 1920x1080 pixels, and a field of view of 69 degrees, and it can capture color images at 30 frames per second.
          The depth sensor in the camera has a resolution of 1280x720 pixels, and a field of view of 58 degrees, and it can capture depth information up to 10 meters.
    \item We added a Pi LIDAR sensor to the robot's base, which is a 2D LIDAR sensor that can capture the distance of the obstacles in the environment.
          The sensor has a range of 12 meters, and a field of view of 360 degrees, and it can capture 4000 samples per second.
    \item We replaced the original processor of the robot with a Jetson Orin Nano board, which delivers up to 40 TOPS of AI performance.
          The sensors are connected to the Jetson board through USB 3.0 ports, and it connects to Pepper via an Ethernet cable.
    \item To power the new hardware, we attached a 11,1V 8,000mAH LiPo battery to the robot's torso and connected it to the Jetson board directly.
          And considering the Jetson maximum power consumption of 15W, and 5W for the sensors, the battery can power the new hardware for more than 4:20' hours.
\end{itemize}
\noindent
Thanks to these upgrades (see Fig.~\ref{fig:Pepper-modified}), the robot can now detect the 3D pose of the people in the environment, and the obstacles around it,
and run social navigation algorithms in real-time.

\vspace{-0.32cm}
\bibliographystyle{IEEEtran}
\bibliography{IEEEabrv,refs}

\end{document}